\def\year{2018}\relax
\documentclass[letterpaper]{article} 
\usepackage{aaai18}  
\usepackage{times}  
\usepackage{helvet}  
\usepackage{courier}  
\usepackage{url}  
\usepackage{graphicx}  
\usepackage{microtype}      
\usepackage{amsmath}
\usepackage{amssymb}
\usepackage{color}
\usepackage{epsfig}
\begin{document}
%
\title{Recurrent Attentional Reinforcement Learning for Multi-label Image Recognition}
\author{Tianshui Chen$^1$, Zhouxia Wang$^{1,2}$, Guanbin Li$^{1}$\thanks{Corresponding author is Guanbin Li (Email: liguanbin@mail.sysu.edu.cn). This work was supported by State Key Development Program under Grant 2016YFB1001004, the National Natural
Science Foundation of China under Grant 61622214 and Grant 61702565, and Special Program of the NSFC-Guangdong Joint Fund for Applied Research on Super Computation (the second phase). This work was also sponsored by CCF-Tencent Open Research Fund.}, Liang Lin$^{1,2}$\\
$^1$School of Data and Computer Science, Sun Yat-sen University, Guangzhou, China\\
$^2$SenseTime Group Limited\\
}
\maketitle

\begin{abstract}
Recognizing multiple labels of images is a fundamental but challenging task in computer vision, and remarkable progress has been attained by localizing semantic-aware image regions and predicting their labels with deep convolutional neural networks. The step of hypothesis regions (region proposals) localization in these existing multi-label image recognition pipelines, however, usually takes redundant computation cost, e.g., generating hundreds of meaningless proposals with non-discriminative information and extracting their features, and the spatial contextual dependency modeling among the localized regions are often ignored or over-simplified. To resolve these issues, this paper proposes a recurrent attention reinforcement learning framework to iteratively discover a sequence of attentional and informative regions that are related to different semantic objects and further predict label scores conditioned on these regions. Besides, our method explicitly models long-term dependencies among these attentional regions that help to capture semantic label co-occurrence and thus facilitate multi-label recognition. Extensive experiments and comparisons on two large-scale benchmarks (i.e., PASCAL VOC and MS-COCO) show that our model achieves superior performance over existing state-of-the-art methods in both performance and efficiency as well as explicitly identifying image-level semantic labels to specific object regions.
\end{abstract}

\section{Introduction}
Image classification, as a foundational problem in computer vision, is receiving increasing attention in the research community. Although marked progress is achieved in this topic thanks to the great success of deep convolutional neural networks (CNNs) \cite{krizhevsky2012imagenet,he2016deep}, existing approaches mainly focus on single-label image classification that considers the situation where an image would contain only one object. In contrast, multi-label image recognition shows more practical significance, as the real-world image is normally annotated with multiple labels and modeling rich semantic information is essential for the task of high-level image understanding.

A straightforward method that extends CNNs to multi-label image recognition is to fine tune the networks pre-trained on single-label classification dataset (e.g.,  ImageNet \cite{russakovsky2015imagenet}) and extract global representation for multi-label recognition \cite{chatfield2014return}. Though being end-to-end trainable, classifiers trained on global image representation may not generalize well to images containing multiple objects with different locations, scales, occlusions, and categories. An alternative way \cite{yang2016exploit,wei2016hcp} is to introduce object proposals that are assumed to contain all possible foreground objects in the image, and aggregate features extracted from all these proposals to incorporate local information for multi-label image recognition. Despite notable improvement compared to global representation, these methods still have many flaws. First, these methods need to extract hundreds of proposal to achieve a high recall but feeding such a large number of proposals to the CNN for classification is extremely time-consuming. Second, an image usually contains only several objects, most of the proposals either provide intensely coarse information of an object or even refer to the same object. In this way, redundant computation and sub-optimal performance are inevitable, especially in complex scenarios. Last but not least, they usually oversimplify the contextual dependencies among foreground objects and thus fail to capture label correlations in images. 

In this paper, inspired by the way that humans continually move fovea from one discriminative object to the next when performing image labeling tasks, we propose an end-to-end trainable recurrent attention reinforcement learning framework to adaptively search the attentional and contextual regions in term of classification. Specifically, our proposed framework consists of a fully convolutional network for extracting deep feature representation, and a recurrent attention-aware module, implemented by an LSTM network, to iteratively locate the class-related regions and predict the label scores over these located regions. At each iteration, it predicts the label scores for the current region and searches an optimal location for the next iteration. Note that by ``remember" the information of the previous iterations, the LSTM can naturally capture contextual dependencies among the attentional regions, which is also a key factor that facilitates multi-label recognition \cite{zhang2016multi}. During training, we formulate it as a sequential decision-making problem, and introduce reinforcement learning similar to previous visual attention models \cite{mnih2014recurrent,ba2014multiple}, where the action is searching the attentional location of each glimpse and performing classification on attentional regions, the state is the features regarding the current regions as well as the information of previous iteration, and the reward measures the classification correctness. In this way, the proposed framework is trained with merely image-level labels in an end-to-end fashion, requiring no explicit object bounding boxes.

To the best of our knowledge, this is the first paper that introduces recurrent attentional mechanism with deep reinforcement learning into generic multi-label image classification. Compared to the recent hypothesis-regions-based multi-label recognition methods, our proposed method not only enjoys better computational efficiency and higher classification accuracy, but also provides a semantic-aware object discovery mechanism based on merely image-level labels. Experimental results on two large-scale benchmarks (PASCAL VOC and MS-COCO) demonstrate the superiority of our proposed method against state-of-the-art algorithms. We also conduct experiments to extensively evaluate and discuss the contribution of the crucial components. 

\begin{figure*}[!t]
   \centering
   \includegraphics[width=0.85\linewidth]{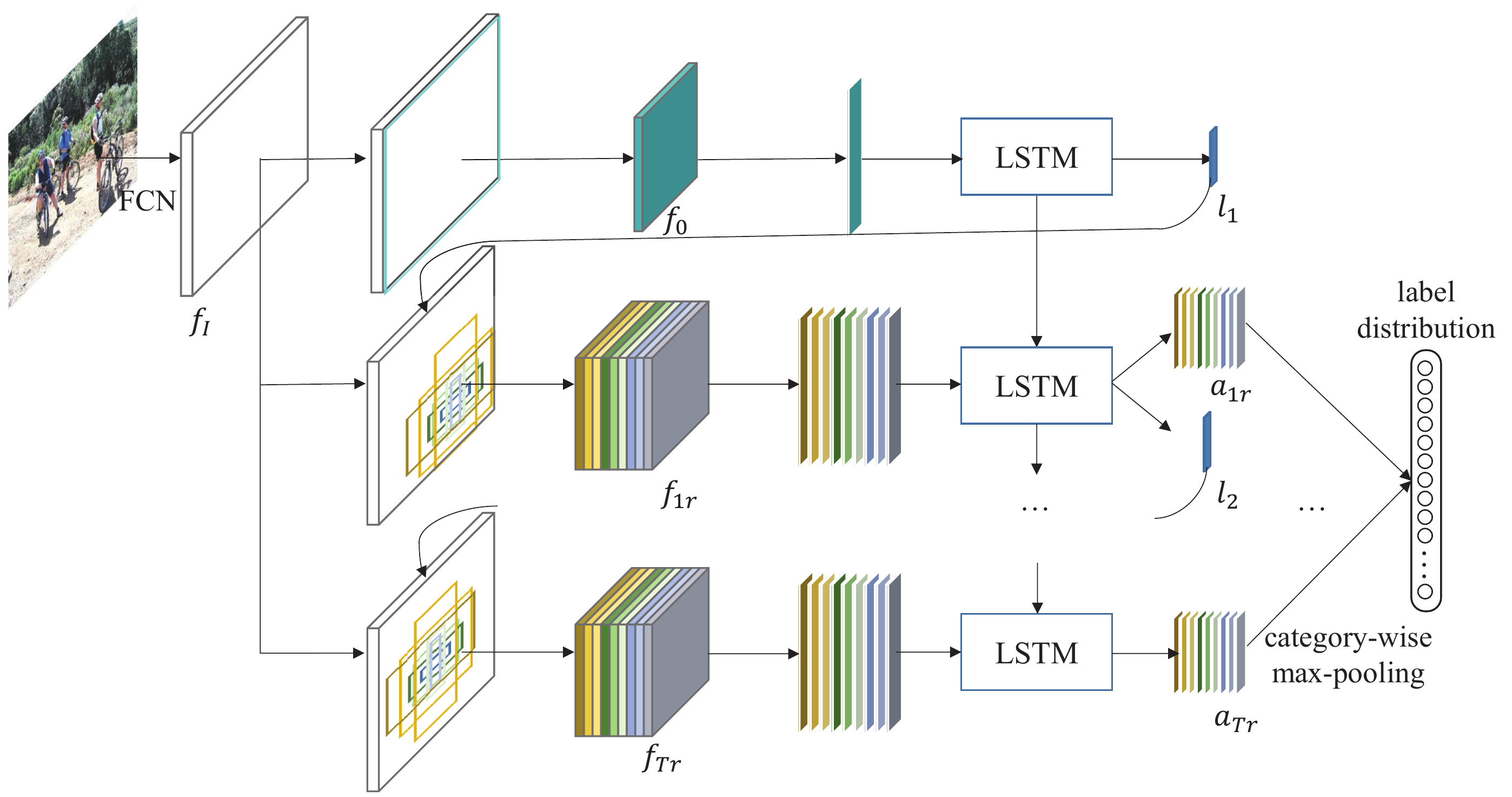}
   \caption{Overview of our proposed framework for multi-label image recognition. The input image is first fed to the VGG16 ConvNet and mapped to the feature maps $f_I$. At each iteration $t$, $k$ regions are yielded at the center location $l_t$ estimated from the previous iteration and corresponding fixed-size features are also extracted. An LSTM unit takes these features as well as the hidden state of the previous iteration as input to predict the scores for each region and searches the location for the next iteration. All the predicted scores are fused using the category-wise max-pooling to obtain the final label distribution. The framework is end-to-end trained using merely image-level labels using reinforcement learning techniques.}
   \label{fig:framework}
\end{figure*}
\section{Related Work}
We review the related works according to two main research streams: multi-label image recognition and visual attention networks.

\subsection{Multi-label image recognition}
Recent progress on single-label image classification is made based on the deep convolutional neural networks (CNNs)\cite{krizhevsky2012imagenet,simonyan2014very,he2016deep} that learn powerful visual representation via stacking multiple nonlinear transformations. Several works have also adapted these single-label classification networks to multi-label image recognition \cite{sharif2014cnn,simonyan2014very,yang2016exploit,wei2016hcp,wang2016cnn}. For example, Razavian et al.~\cite{sharif2014cnn} extract off-the-shelf features using Alex-Net pre-trained on the ImageNet dataset and train an SVM classifier for each category. Chatfield et al.~\cite{chatfield2014return} fine-tune the network using the target multi-label dataset to learn more domain-specific features, which helps to boost the classification accuracy. Gong et al.~\cite{gong2013deep} explore training the CNN via various losses to tackle this problem and comes to a conclusion that the weighted approximate ranking loss can return the best performance. However, these methods treat the labels in the image independently and fail to capture semantic label co-occurrence. Instead, there are a series of works resorting to graphical models to capture pairwise label correlations, including Conditional Random Field~\cite{ghamrawi2005collective}, Dependency Network~\cite{guo2011multi}, and co-occurrence matrix~\cite{xue2011correlative}. Most recently, Wang et al.~\cite{wang2016cnn} formulate a CNN-RNN framework to jointly characterize the semantic label dependency and the image-label relevance. Zhu et al. \cite{zhu2017learning} further propose a Spatial Regularization Network that generates class-related attention maps and captures both spatial and semantic label dependencies via simple learnable convolutions.

The works mentioned above mainly consider the global representation of the whole image. However, a classifier trained using the global representation may not be optimal for multi-label image recognition, since they not only ignore the relationship between semantic labels and local image regions but also are vulnerable to the non-informative background. To address this problem, recent works~\cite{yang2016exploit,wei2016hcp} extract object proposals as the informative regions and aggregate local features on these regions for multi-label recognition. More concretely, Wei et al.~\cite{wei2016hcp} propose a Hypotheses-CNN-Pooling framework, which makes predictions on each proposal and then aggregates all the predictions as the final output through category-wise max-pooling. Yang et al.~\cite{yang2016exploit} formulates the multi-label image recognition as a multi-class multi-instance problem and incorporates feature as well as label view information of the proposals for feature enhancement. Newest work \cite{zhang2016multi} also utilizes CNN-based proposals and simultaneously models label dependencies among the proposals. Despite jointly training the proposals and image classification, this method needs to additionally train the proposal generation component with the annotation of bounding boxes. 

\subsection{Visual attention networks}
One drawback of the proposal-based methods is the necessary for extracting object proposals, preventing the model from end-to-end training \cite{wei2016hcp,yang2016exploit} or requiring extra annotations of the bounding boxes \cite{zhang2016multi}. Recently, visual attention networks have been intensively proposed to automatically mine the relevant and informative regions, which have benefited a broad range of vision tasks, including image recognition ~\cite{mnih2014recurrent,ba2014multiple,xiao2015application}, image captioning ~\cite{xu2015show} and visual question answering~\cite{xiong2016dynamic}. These works usually design a recurrent neural network to iteratively search the attentional regions, which can be formulated as a sequential decision-making problem. Reinforcement learning technique is commonly introduced to optimize the sequential model with delayed reward. Specifically,~\cite{mnih2014recurrent,ba2014multiple} propose a recurrent attention model trained with reinforcement learning to attend the most relevant regions of the input image and demonstrate both accurate and efficient results on the digital classification task. However, it may not generalize well to multi-label classification for generic images as they are far more complex, and different objects undergo drastic changes in both scales and shapes. In this paper, we introduce the recurrent attention mechanism into generic multi-label image classification for locating attentional and contextual regions regarding classification and demonstrate it can still improve multi-label recognition in both accuracy and efficiency.

\section{Proposed Method}
Figure \ref{fig:framework} illustrates an overview of the proposed method. Given an input image $I$, it is first resized to $W\times{H}$ and fed into the VGG16 ConvNet \cite{simonyan2014very}. The ConvNet processes on the whole image with multiple stacked convolutional layers to produce the feature maps $f_I\in{\mathcal{R}^{C\times{W'}\times{H'}}}$. Here, we use the feature maps from the last convolutional layer (i.e., conv5\_3). The core of our proposed method is the recurrent attention-aware module that locates the attentional regions and predicts the label scores for these regions in an iterative manner. Finally, the scores over all attentional regions are fused to get the final label distribution. In the following context, we introduce this module in detail.

At iteration $t$, the agent first receives a location $l_t$ computed at the previous iteration and extracts regions based on $l_t$. Previous works \cite{mnih2014recurrent} simply extract square patches centered at $l_t$. However, general objects undergo drastic changes in both shapes and scales, and thus directly extracting squre patches can hardly cover all these objects. Inspired by the anchor strategy proposed in \cite{ren2015faster}, we yield $k$ regions $R_t=\{R_{tr}\}_{r=1}^k$ related to various scales and aspect ratios centered at $l_t$ and then extract the features for all these regions:
\begin{equation}
   f_{tr}=\mathcal{G}(f_I,R_{tr}), r=1,2,\cdots,k,
\end{equation}
where $\mathcal{G}$ consists of a cropping operation that crops the region $R_{tr}$ on $f_I$, followed by a bilinear interpolation that maps the cropped feature maps to a fixed-size counterpart $f_{tr}$. Previous works \cite{mnih2014recurrent} crop the regions at original input image and apply CNN for repeatedly extracting features for each region, leading to a high computational burden. Instead, we apply the operation on the feature maps $f_I$ to avoid repeating the convolutional processes that are computationally intensive, significantly improving the efficiency during both training and test stages. Once the features are extracted, the recurrent attention-aware module, implemented by an LSTM network \cite{hochreiter1997long}, takes the hidden state of the previous iteration as well as the features of currently located regions as input, predicts the classification scores for each region and searches an optimal location for the next iteration, formulated as:
\begin{equation}
   \{a_{t1},a_{t2},\cdots,a_{tk},l_{t+1}\}=\mathcal{T}_{\pi}(f_{t1},f_{t2},\cdots,f_{tk},h_{t-1};\theta)
\end{equation}
where $\mathcal{T}_{\pi}(\cdot)$ represents the recurrent attention-aware module, and $\theta$ denotes the network parameters. $a_{tr}$ is the label score vector with respect to the region $R_{tr}$. The initial region is set as the whole image, so $R_0$ has only one region, and it is merely used to determine the location $l_1$.

\noindent\textbf{Category-wise max-pooling.} The iterations are repeated for $T+1$ times, yielding $T\times{k}$ label score vectors, i.e., $\{a_{tr}|t=1,2,\cdots,T; r=1,2,\cdots,k\}$, where $a_{tr}=\{a_{tr}^{0},a_{tr}^{1},\cdots,a_{tr}^{C-1}\}$ is the score vector of region $R_{tr}$ over $C$ class labels. Following previous work \cite{wei2016hcp}, we utilize the category-wise max-pooling operation to fuse these score vectors and obtain the final result $a=\{a^0,a^1,\cdots,a^{C-1}\}$ via simply maximizing out the scores over regions for each category, formulated as:

 \begin{equation}
      a^c=\mathrm{max}(a_{11}^c,a_{12}^c,\dots,a_{Tk}^c), c=0,1,\dots,C-1.
\end{equation}

\subsection{Recurrent attention-aware module}
The recurrent attention-aware module iteratively predicts the label scores of the current regions and searches a most relevant location for the next iteration, which can be regarded as a sequential decision-making problem. At each iteration, it takes action to predict the label scores for the attended regions and searches an optimal location conditioned on the current states. After the action, the state is updated by a new hidden state and a newly attended location. The process is repeated until a maximum iteration is reached. In the end, the scores of all the located regions are fused to get the final label distribution, and a delayed global reward, which is computed based on this predicted result and the ground-truth labels, is employed to guide the agent training. We elaborate the involved states, actions and reward signal in the following.

\noindent\textbf{State.} The state $s_t$ should provide sufficient information for the agent to make decisions. Concretely, it should encode the knowledge of the current environment and those of the previous iterations. To this end, it comprises two parts: 1) the features of the current regions (i.e., $\{f_{tr}\}_{r=1}^k$), which is instrumental to classification and provides rich contextual information to help the agent to mine more complemental and discriminative regions; 2) the hidden state of the previous iteration $h_{t-1}$, which encodes the information of the past iterations and updates over time via the LSTM module. Moreover, simultaneously considering the information of the previous iterations can also help to capture the contextual dependencies among all the glimpsed regions and labels. In this way, by sequentially observing the states $s_t=\{f_{t1}, f_{t2}, \cdots, f_{tk}, h_{t-1}\}$, the agent is capable of performing classification for the current regions and determining the next optimal location.

\noindent\textbf{Action.} Given the state $s_t$, the agent takes two actions: 1) performing classification on the current attentional regions; 2) searching an optimal location $l_{t+1}$ over all possible locations $\{l_{t+1}=(x,y)|0\le{x}\le{W'},0\le{y}\le{H'}\}$ on the feature map $f_I$. As shown in figure \ref{fig:framework}, a fully-connected layer is utilized to map the extracted features $f_{tr}$ to the semantic representation for each attended region. The LSTM unit takes the semantic representation and the hidden state of the previous iteration as input, and produces a new hidden state $h_{tr}$. Finally, the classification scores can be computed through a small classification network, denoted as:

 \begin{equation}
      a_{tr}=f_{cls}(h_{tr};\theta_{cls}), r=1,2,\cdots,k,
\end{equation}
where the classification network $f_{cls}(\cdot)$ is implemented by a fully-connected layer, with $\theta_{cls}$ being its parameters. For the localization action, all the hidden states are first averaged to get a final hidden state, denoted as $h_t=\frac{1}{k}\sum_r{h_{tr}}$. Then the agent builds a gaussian distribution $P(l_{t+1}|f_{loc}(h_t;\theta_{loc}), \sigma)$. In this equation, $f_{loc}(h_t;\theta_{loc})$, the localization network output, is set as the mean value of the distribution, and $\sigma$ is its standard deviation and is empirically set as 0.11. Similarly, the localization network $f_{loc}(\cdot)$ is also implemented by a fully-connected layer parameterized by $\theta_{loc}$. At iteration $t$, the agent selects the localization action $l_{t+1}$ by randomly drawing a location over the probability distribution.

\noindent\textbf{Reward.}  After executing the actions at each iteration, the agent updates the state and receives a reward signal. For the task of multi-label image recognition, it is desired to aggregate the predictions over all located regions for counting the reward, since each region is expected to be associated with one semantic label. Thus, we define a delayed reward assignment mechanism based on the final aggregated result. For a sample with $n$ ground-truth labels, its label set is $g=\{l_1^g, l_2^g, \cdots, l_n^g\}$. We then sort the predicted scores and obtain the predicted label set $p=\{l_1^p, l_2^p, \cdots, l_n^p\}$ with top-$n$ scores. The reward at iteration $t$ is defined as:
\begin{equation}
    r_t=
   \begin{cases}
    \frac{|g\cap{p}|}{n} &\mbox{$t=T$}\\
    0 &\mbox{$t<T$}
   \end{cases}
\end{equation}
where $|\cdot|$ is the cardinality of the set. We aim to maximize the sum of the discounted rewards:
 \begin{equation}
      R=\sum_{t=1}^T{\gamma^{t-1}r_t}
\end{equation}
where $\gamma$ is the discount factor. We set $\gamma$ as 1 in our experiments, and the total reward is $R=r_T$. In this work, we utilize the reward to guide the agent to search the optimal actions.

\subsection{Optimization}
At the training stage, in addition to defining the similar classification loss with \cite{yang2016exploit}, we take the delayed reward assignment into account for optimizing the region localization policy, leading to a hybrid objective function for model training. In the experiments, the model is trained with the hybrid loss in an end-to-end manner.

Formally, the agent needs to learn a policy $\pi((a_t,l_{t+1})|S_t;\theta)$, which predicts a distribution over actions for the current iteration based on the sequence of past observations and actions taken by the agent, i.e.,  $S_t=R_0,l_1,R_1,a_1,l_2,\cdots,R_t$. To this end, we define the objective function to maximize the expectation of the reward, expressed as:
\begin{equation}
      \mathcal{J}(\theta)=\mathbb{E}_{P(S_T;\theta)}[R].
\end{equation}
where $P(S_T;\theta)$ is the distribution over all possible interaction sequences, and it is dependent on the policy. Inspired by the work ~\cite{mnih2014recurrent}, we leverage the REINFORCE algorithm~\cite{williams1992simple} from the reinforcement learning community to estimate the gradient for backpropagation. Specifically, it utilizes sample approximation to compute the gradients, formulated as:
\begin{equation}
   \begin{split}
      \nabla{\mathcal{J}(\theta)}&=\sum_{t=1}^T\mathbb{E}_{P(S_T;\theta)}[\nabla_{\theta}\log{\pi((a_t,l_{t+1})|S_t;\theta)}R]\\
      &\approx \frac{1}{M}\sum_{i=1}^M\sum_{t=1}^T [\nabla_{\theta}\log{\pi((a_t^i,l_{t+1}^i)|S^i_t;\theta)}R^i]
         \end{split}
         \label{RL_obj}
\end{equation}
where $i=1,2,\cdots,M$ denotes the index of the $M$ episodes. However, the gradient estimated using Equation (\ref{RL_obj}) is of high variance, and consequently, the training process is difficult to converge. To solve this problem, we further employ the variance reduction strategy proposed in~\cite{mnih2014recurrent} to obtain an unbiased low-variance gradient estimation.

The policy is learnt using the delayed reward signal, as the ``best'' action at each iteration is unavailable. In the context of multi-label recognition, the ground-truth labels for each sample exist. Thus, we further define a loss function following \cite{wei2016hcp,yang2016exploit} as the extra supervision. Suppose there are $N$ training samples, and each sample $x_i$ has its label vector $y_i=\{y_{i}^{0},y_{i}^{1},\dots,y_{i}^{C-1}\}$. $y_{i}^c$ $(c=0,1,\dots,C-1)$ is assigned as 1 if the sample is annotated with the class label $c$, and 0 otherwise. The ground-truth probability vector of the $i$-th sample is defined as $\hat{p}_i=y_i/||y_i||_1$, and the classification loss function is thus formulated as:
 \begin{equation}
      \mathcal{L}_{\textrm{cls}}=\frac{1}{N}\sum_{i=1}^N\sum_{c=0}^{C-1}(p_{i}^c-\hat{p}_{i}^c)^2.
\end{equation}
where $p_i$ is the predicted probability vector and can be computed via:
 \begin{equation}
      p_{i}^c= \frac{\exp(a_{i}^c)}{\sum_{c'=0}^{C-1}\exp(a_{i}^{c'})} \ c=0,1,\dots,C-1.
\end{equation}

\section{Experiments}
In this section, we present extensive experimental results and comparisons that demonstrate the superiority of the proposed method. We also conduct experiments to carefully evaluate and discuss the contribution of the crucial components.

\subsection{Experiment setting}
\subsubsection{Implementation details} 
During training, all the images are resized to $N\times{N}$, and randomly cropped with a size of $(N-64)\times(N-64)$, followed by a randomly horizontal flipping, for data augmentation. In our experiments, we train two models with $N=512$ and $N=640$, respectively. For the anchor strategy, we set 3 region scales with area $80\times{80}$, $160\times{160}$, $320\times{320}$ for $N=512$ and $100\times{100}$, $200\times{200}$, $400\times{400}$ for $N=640$, and 3 aspect ratios of 2:1, 1:1, 1:2 for both scales. Thus, $k$ is set as 9. Both of the models are optimized using the Adam solver with a batch size of 16, an initial learning rate of 0.00001, momentums of 0.9 and 0.999. During testing, we follow~\cite{krizhevsky2012imagenet} to perform ten-view evaluation across the two scales. Specifically, we first resize the input image to $N\times{N}$ ($N=512, 640$), and extract five patches (i.e., the four corner patches and the center patch) with a size of $(N-64)\times(N-64)$, as well as their horizontally flipped counterparts. In the experiments, instead of repeatedly extracting features for each patch, we feed the $N\times{N}$ image to the VGG16 ConvNet and crop the features on the conv5\_3 features maps accordingly for each patch. In this way, the computational complexity is remarkably reduced. The model predicts a label score vector for each view, and the final result is computed as the average predictions over the ten views.

\subsubsection{Evaluation metrics} 
We first employ the average precision (AP) for each category, and the mean average precision (mAP) over all categories to evaluate all the methods. We also follow \cite{gong2013deep,wang2016cnn} to compute the precision and recall for the predicted labels. For each image, we assign top $k$ highest-ranked labels to the image, and compare with the ground-truth labels. The precision is the fraction of the number of the correctly predicted labels in relation to the
number of predicted labels; The recall is the fraction of the number of the correctly predicted labels in relation to the number of ground-truth labels. In the experiments, we compute the overall precision, recall, $F1$ ($OP$, $OR$, $OF1$) and
per-class precision, recall, $F1$ ($CP$, $CR$, $CF1$) for comparison, which can be computed as: 
\begin{equation}
   \begin{split}
         OP&=\frac{\sum_{i}N_{i}^{c}}{\sum_{i}N_{i}^{p}},\quad\\
         OR&=\frac{\sum_{i}N_{i}^{c}}{\sum_{i}N_{i}^{g}},\quad\\
         OF1&=\frac{2 \times OP \times OR}{OP+OR},\quad 
   \end{split}
      \begin{split}
         CP&=\frac{1}{C}\sum_{i}\frac{N_{i}^{c}}{N_{i}^{p}}\\
         CR&=\frac{1}{C}\sum_{i}\frac{N_{i}^{c}}{N_{i}^{g}}\\
         CF1&=\frac{2 \times CP \times CR}{CP+CR}
   \end{split}
   \label{eqn:metric}
\end{equation}
where $C$ is the number of labels, $N_{i}^{c}$ is the number of images that are correctly predicted for the $i$-th label, $N_{i}^{p}$ is the number of predicted images for the $i$-th label, $N_{i}^{g}$ is the number of ground truth images for the $i$-th label.

\begin{table*}[htp]
\centering
\tiny
\begin{tabular}{c|cccccccccccccccccccc|c}
\hline
\centering Methods  & aero & bike & bird & boat & bottle & bus & car & cat & chair & cow & table & dog & horse & mbike & person & plant & sheep & sofa & train & tv & mAP \\
\hline
\hline
\centering CNN-SVM & 88.5 & 81.0  & 83.5 &  82.0 & 42.0 & 72.5 & 85.3 & 81.6 & 59.9 & 58.5 & 66.5 & 77.8 & 81.8 & 78.8 & 90.2 & 54.8 & 71.1 & 62.6 & 87.2  & 71.8 & 73.9 \\
\centering CNN-RNN & 96.7 & 83.1 & 94.2 & 92.8 & 61.2 & 82.1 & 89.1 & 94.2 & 64.2 & 83.6 & 70.0 & 92.4 & 91.7 & 84.2 & 93.7 & 59.8 & 93.2 & 75.3 & \textcolor[rgb]{1,0,0}{99.7} & 78.6 & 84.0 \\
\centering VeryDeep  & \textcolor[rgb]{1,0,0}{98.9} & 95.0 & 96.8 & 95.4 & 69.7 & 90.4 & 93.5 & 96.0 & 74.2 & 86.6 & \textcolor[rgb]{1,0,0}{87.8} & 96.0 & 96.3 & 93.1 & 97.2 & 70.0 & 92.1 & 80.3 & 98.1 & 87.0 & 89.7\\
\centering RLSD & 96.4 &  92.7 & 93.8 & 94.1 & 71.2 &  92.5 &  94.2 & 95.7 & 74.3 &  90.0 & 74.2  & 95.4 & 96.2 &  92.1 & 97.9 & 66.9 & \textcolor[rgb]{0,0,1}{93.5} & 73.7 & 97.5 & 87.6 & 88.5\\
\centering HCP  & \textcolor[rgb]{0,0,1}{98.6} & \textcolor[rgb]{1,0,0}{97.1}  & \textcolor[rgb]{1,0,0}{98.0} & \textcolor[rgb]{1,0,0}{95.6} & \textcolor[rgb]{0,0,1}{75.3}&{\noindent\color{red}{94.7}}  & 95.8 &\textcolor[rgb]{1,0,0}{97.3} & 73.1 & 90.2 & 80.0 & \textcolor[rgb]{1,0,0}{97.3} & 96.1 & \textcolor[rgb]{0,0,1}{94.9} & 96.3 & 78.3 & \textcolor[rgb]{1,0,0}{94.7} & 76.2 & 97.9 & \textcolor[rgb]{1,0,0}{91.5} & 90.9\\
\centering FeV+LV& 97.9 & \textcolor[rgb]{0,0,1}{97.0} & 96.6 & 94.6 & 73.6 & \textcolor[rgb]{0,0,1}{93.9} & \textcolor[rgb]{0,0,1}{96.5}& 95.5 & 73.7 & 90.3 & 82.8 & 95.4 & \textcolor[rgb]{1,0,0}{97.7} & \textcolor[rgb]{1,0,0}{95.9} & \textcolor[rgb]{1,0,0}{98.6} & 77.6 & 88.7 & 78.0 & 98.3 & 89.0 & 90.6\\
\hline
\centering Ours (512)  & \textcolor[rgb]{0,0,1}{98.6} & 96.9 & 96.3 & 94.8 & 74.1 & 91.9 & 96.3 & \textcolor[rgb]{0,0,1}{97.1} & \textcolor[rgb]{0,0,1}{76.9} & \textcolor[rgb]{0,0,1}{91.4} & 86.2 & 96.6 & 96.4 & 93.1 & 98.0 & 79.8 & 91.7 & \textcolor[rgb]{0,0,1}{83.1} & 98.3 & 88.6  & \textcolor[rgb]{0,0,1}{91.3} \\
 \centering Ours (640) & 97.9 & \textcolor[rgb]{1,0,0}{97.1} & 96.9 & 95.3 & \textcolor[rgb]{0,0,1}{75.3} & 91.8 & \textcolor[rgb]{0,0,1}{96.5} & 96.7 & 76.8 & 91.0 & 85.6 & 95.7 & 96.0 & 93.5 & 98.2 & \textcolor[rgb]{0,0,1}{81.0} & 92.7 & 80.6 & 98.2 & 89.0 & \textcolor[rgb]{0,0,1}{91.3} \\
\centering Ours          &\textcolor[rgb]{0,0,1}{98.6} & \textcolor[rgb]{1,0,0}{97.1} & \textcolor[rgb]{0,0,1}{97.1} & \textcolor[rgb]{0,0,1}{95.5} & \textcolor[rgb]{1,0,0}{75.6} & 92.8 & \textcolor[rgb]{1,0,0}{96.8} & \textcolor[rgb]{1,0,0}{97.3} & \textcolor[rgb]{1,0,0}{78.3} & \textcolor[rgb]{1,0,0}{92.2} & \textcolor[rgb]{0,0,1}{87.6} & \textcolor[rgb]{0,0,1}{96.9} & \textcolor[rgb]{0,0,1}{96.5} & 93.6 &\textcolor[rgb]{0,0,1}{98.5} & \textcolor[rgb]{1,0,0}{81.6} & 93.1 & \textcolor[rgb]{1,0,0}{83.2} & \textcolor[rgb]{0,0,1}{98.5} & \textcolor[rgb]{0,0,1}{89.3} & \textcolor[rgb]{1,0,0}{92.0} \\
\hline
\end{tabular}
\caption{ Comparison results of AP and mAP in \% of our model and the previous state of the art methods on the VOC07 dataset. The best results and second best results are highlighted in {\color{red}{red}} and {\color{blue}{blue}}, respectively. Best viewed in color.}
\label{table:comparision_voc07}
\end{table*}

\subsection{Comparison with state-of-the-art methods}
To prove the effectiveness of the proposed method, we conduct comprehensive experiments on two widely used benchmarks: Pascal VOC 2007 (VOC07) \cite{everingham2010pascal} and Microsoft COCO (MS-COCO) \cite{lin2014microsoft}.

\subsubsection{Performance on the VOC07 dataset}
The VOC07 dataset contains 9,963 images of 20 object categories, and it is divided into trainval and test sets. It is the most widely used benchmark for multi-label image recognition, and most competing methods have reported their results on this dataset. We compare our model against the following state-of-the-art methods: FeV+LV~\cite{yang2016exploit}, HCP~\cite{wei2016hcp}, CNN-RNN~\cite{wang2016cnn}, RLSD \cite{zhang2016multi}, VeryDeep~\cite{simonyan2014very} and CNN-SVM~\cite{sharif2014cnn}. Note that we report the results of FeV+LV and HCP using VGG-16 ConvNet for fair comparisons. Following the competitors, we train our model on the trainval set and evaluate the performance on the test set.

The comparison results are summarized in Table~\ref{table:comparision_voc07}. As shown, the previous best-performing methods are HCP and FeV+LV, both of which extract hundreds of object proposals, and then aggregate the features of these object proposals for multi-label recognition. They achieve mAPs of 90.9\% and 90.6\%, respectively. Different from these two methods, our model learns an optimal policy to locate a sequence of discriminative regions, while simultaneously trains classifiers to perform classification on these attended regions. In this way, our model can better explore the relations between semantic labels and attentional regions, leading to the performance improvement. Specifically, our model achieves a mAP of 92.0\%, suppressing all the previous state-of-the-art methods by a sizable margin. It is noteworthy that the performance with one single scale of 512 or 640 also performs better than existing methods, further demonstrating the superiority of our model.

\begin{table}[htp]
\centering
\begin{tabular}{c|ccc|ccc}
\hline
\centering Methods  & C-P & C-R & C-F1 & O-P & O-R & O-F1  \\
\hline
\hline
\centering  WARP  & 59.3 & 52.5 & 55.7 & 59.8 & 61.4 & 60.7   \\
\centering  CNN-RNN & 66.0 & 55.6 & 60.4 & 69.2 & \textcolor[rgb]{1,0,0}{66.4} & 67.8 \\
\centering RLSD  & 67.6 &  57.2 &  62.0 & 70.1 & \textcolor[rgb]{0,0,1}{63.4} & 66.5 \\
\hline
\centering  Ours (512)  & 77.5 &  \textcolor[rgb]{0,0,1}{56.8} &  \textcolor[rgb]{0,0,1}{65.6} & 83.0 & 61.2 & \textcolor[rgb]{0,0,1}{70.5} \\
\centering  Ours (640)  & \textcolor[rgb]{0,0,1}{77.9} & 56.3 & 65.4 &  \textcolor[rgb]{0,0,1}{83.5} & 61.0 &  \textcolor[rgb]{0,0,1}{70.5} \\
\centering  Ours & \textcolor[rgb]{1,0,0}{78.8} &\textcolor[rgb]{1,0,0}{57.2} & \textcolor[rgb]{1,0,0}{66.2} & \textcolor[rgb]{1,0,0}{84.0} &  61.6 & \textcolor[rgb]{1,0,0}{71.1} \\
\hline
\end{tabular}
\caption{Comparison results of our model and the previous state of the art methods on the MS-COCO dataset. The best and second best results are highlighted in {\color{red}{red}} and {\color{blue}{blue}}, respectively. Best viewed in color.}
\label{table:coco_comparison}
\end{table}

\subsubsection{Performance on the MS-COCO dataset}
The MS-COCO dataset is originally built for object detection and has also been used for multi-label recognition recently. It is a larger and more challenging dataset, which comprises a training set of 82,081 images and a validation set of 40,137 images from 80 object categories. We compare our model with three state-of-the-art methods, i.e., CNN-RNN~\cite{wang2016cnn}, RLSD \cite{zhang2016multi} and WARP~\cite{gong2013deep}, on this dataset. Our method and all the competitors are trained on the train set and evaluated on the validation set since the ground truth labels of the test set are unavailable. Following~\cite{wang2016cnn}, when computing the precision recall metrics, we select the top 3 labels for each image. We also filter out the labels with probabilities lower than a pre-defined threshold (0.1 in our experiments), so the label number of some images would be less than 3.

The comparison results of the overall precision, recall, $F1$, per-class precision, recall, $F1$ are reported in Table~\ref{table:coco_comparison}. Our model significantly outperforms previous methods. Specifically, it achieves a per-class $F1$ score of 66.2\%, an overall $F1$ score of 71.1\%, beating the previous best method by 4.2\% and 3.3\%, respectively.  Similarly, the performance using single scale is still higher than those of other methods.

\subsection{Ablation Study}
In this subsection, we perform ablative studies to carefully evaluate and discuss the contribution of the critical components of our proposed model.

\subsubsection{Effectiveness of the attentional regions} 
The key component of our method is the recurrent attention-aware module that automatically locates the discriminative regions. In this part, we further implement two baseline methods to verify the effectiveness of the attentional regions. The first method replaces the locations attended by our model with randomly selected locations, and it also utilizes 9 anchors for each location. The second method utilizes the representative object proposals as the informative regions to replace the attended regions for classification. This method first employs EdgeBox \cite{zitnick2014edge} to extract proposals and adopts non-maximum suppression with a threshold of 0.7 on them based on their objectness scores to exclude the seriously overlapped proposals. The proposals with the top 5 scores are selected. Table \ref{table:attention} presents the comparison results. Our attentional model evidently outperforms these two baseline methods.

\begin{table}[htp]
\centering
\begin{tabular}{c|c}
\hline
\centering Method & mAP (\%)  \\
\hline 
\hline
\centering random  & 89.0 \\
\centering proposal  & 88.6 \\
\centering attention & 90.2 \\
\hline
\end{tabular}
\caption{Comparison of mAP in \% of our model with attentional regions, proposals, and random regions on the VOC07 dataset. The results are evaluated using single-view with the scale of $512\times 512$.}
\label{table:attention}
\end{table}

\subsubsection{Significance of adopting the LSTM}
To demonstrate the significance of adopting the LSTM, we have conducted two experiments and reported the results in Table \ref{table:lstm}. First, we use the LSTM to locate the regions while removing the classification branch and independently classifying the regions by designing a network, obtaining a lower mAP of 90.0\%. We further remove the LSTM and also predict the locations independently, obtaining an even lower mAP of 89.6\%. These results show the contextual dependencies among the attentional regions captured by the LSTM is crucial for improving the region localization accuracy as well as the multi-label classification accuracy.

\begin{table}[htp]
\centering
\begin{tabular}{c|c}
\hline
\centering method  & mAP (\%) \\
\hline 
\hline
\centering Ours-A & 89.6  \\
\centering Ours-B  & 90.0  \\
\centering Ours-C & 91.3 \\
\hline
\end{tabular}
\caption{Comparison of mAP in \% of using different methods for localization and classification. We report the results using LSTM for both classification and localization (Ours-C), using LSTM for localization but not for classification (Ours-B), and not using LSTM (Ours-A). The results are evaluated using ten-view with the scale of $512\times 512$.}
\label{table:lstm}
\end{table}

\subsubsection{Effectiveness of multiple regions with variable scales and aspect ratios} 
\begin{figure}[htp]
   \centering
   \includegraphics[width=0.98\linewidth]{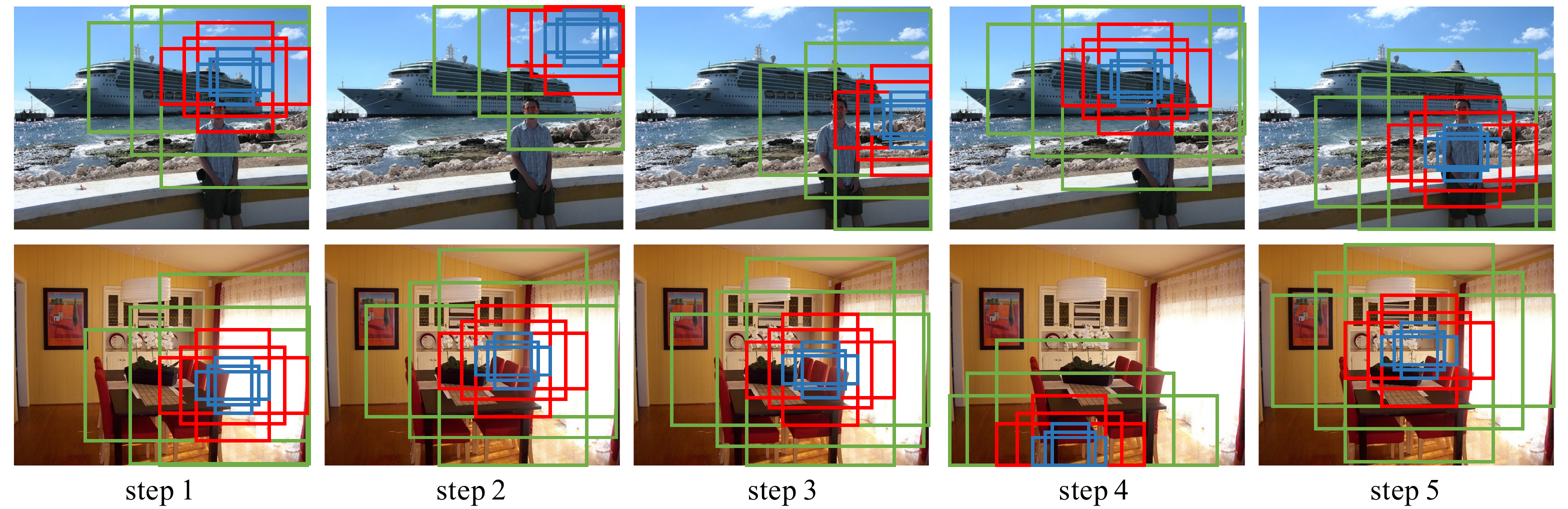}
   \caption{Visualization of the located regions at each iteration. The attentional regions can well locate most semantic objects in the images.}
   \label{fig:visualization}
\end{figure}

As general objects vary dramatically in scale and aspect ratio, we extract 9 regions with 3 scales and 3 aspect ratios, at each iteration. We first visualize some examples of the located regions at each iteration in Figure \ref{fig:visualization}. As shown, different regions can indeed find objects with different scales and aspect ratios. For example, the first image in Figure~\ref{fig:visualization} contains a man and a boat, which vary significantly in scale and aspect ratio. However, both of them can be well located. Concretely, the region with the largest scale and ratio of 2:1 well locates the boat at iteration 4, while the region with the middle scale and ratio of 1:2 can catch the man at iteration 5. The located regions of the second image also exhibit similar results. 

To clearly validate its advantage, we further conduct an experiment that extracts one single region at each location and re-trains the model for comparison. Specifically, we utilize the input image scale of $512\times512$, and for each predicted location, one region with the size of $224 \times 224$ is extracted. The comparison results are depicted in Table~\ref{table:anchor}. It shows that using multiple regions at each iteration leads to better classification performance. 

\begin{table}[htp]
\centering
\begin{tabular}{c|c}
\hline
\centering Method & mAP (\%)  \\
\hline 
\hline
\centering multiple regions & 91.3 \\
\centering single region & 90.9 \\
\hline
\end{tabular}
\caption{Comparison of mAP in \% of our model that extract multiple and single regions at each iteration on the VOC07 dataset. The results are evaluated using ten-view with the scale of $512\times 512$.}
\label{table:anchor}
\end{table}

\subsubsection{Analysis of increasing the recursive iteration}
In this part, we explore the effect of using different recursive iterations $T$. To this, we train our model with different iterations, i.e., $T=1,5,10$, and report the experimental results in Table~\ref{table:recursive_step}. When the iteration increases from 1 to 5, the performance has a notable improvement since the located regions may cover more discriminative objects. However, when further increasing the iteration number, the performance does not improve. One possible reason is that when the iteration is greater than 5, the agent has almost mined out all discriminative regions, and locating more regions make litter sense or even bring noise and redundant computation. Thus, in our experiments, the iteration number is set as 5 to better balance the efficiency and effectiveness.

\begin{table}[htp]
\centering
\begin{tabular}{c|c}
\hline
\centering $T$  & mAP (\%) \\
\hline 
\hline
\centering 1 & 90.9  \\
\centering 5  & 91.3  \\
\centering 10 & 91.3 \\
\hline
\end{tabular}
\caption{Comparison of mAP in \% of our model using different recursive iterations on the VOC07 dataset. The results are evaluated using ten-view with the scale of $512\times 512$.}
\label{table:recursive_step}
\end{table}

\subsection{Efficiency analysis}
Efficiency is another important metric for the real-world systems. In this part, we analyze the execution time of our model and the previous state-of-the-art methods. We test our model on a desktop with a single NVIDIA GeForce GTX TITAN-X GPU. It takes about 150ms for ten-view evaluation for scale 512 and about 200 ms for scale 640. Thus, the execution time of our method is about 350ms per image. However, recent proposal-based methods, e.g., HCP \cite{wei2016hcp} and FeV+LV \cite{yang2016exploit}, need to compute the proposals and repeat processing hundreds of proposals using the deep CNNs, rendering them extremely inefficient. As shown in ~\cite{wei2016hcp}, these methods may take about 10s to process an image on a similar GPU environment, about 30$\times$ slower than ours. 

\section{Conclusion}
In this paper, we propose a recurrent attention reinforcement learning framework that is capable of automatically locating the attentional and informative regions regarding classification, and predicts the label scores over all attentional regions. We formulate the region localization process as a sequential decision-making problem and resort to reinforcement learning technique to optimize the proposed framework with merely image-level labels in an end-to-end manner. Compared to the previous proposal-based methods, our method can better explore the interaction between semantic labels and attentional regions, while explicitly capturing the contextual dependencies among these regions. Extensive experimental results and evaluations on two large-scale and challenging benchmarks, i.e., Pascal VOC and MicroSoft COCO, well demonstrate the superiority of our proposed method on both accuracy and efficiency. 


\bibliographystyle{aaai}
\bibliography{reference}


\end{document}